\documentclass[letterpaper]{article} 
\usepackage{aaai25}  
\usepackage{times}  
\usepackage{helvet}  
\usepackage{courier}  
\usepackage[hyphens]{url}  
\usepackage{graphicx} 
\usepackage{natbib}  
\usepackage{caption} 
\usepackage{algorithm}
\usepackage{algorithmic}

\usepackage[table]{xcolor}

\usepackage{color}
\usepackage{booktabs} 
 \usepackage{multirow}
\usepackage{colortbl}
\usepackage{cite}
\usepackage{bm}
\usepackage{amsmath,amsfonts}
\usepackage[bottom]{footmisc}

\newcommand{\argmin}{\arg\!\min}

\newcommand{\Amat}{{\boldsymbol A}}
\newcommand{\Bmat}{{\boldsymbol B}}
\newcommand{\Cmat}{{\boldsymbol C}}
\newcommand{\Dmat}{{\boldsymbol D}}

\newcommand{\Imat}{{\boldsymbol I}}

\newcommand{\Xmat}{{\boldsymbol X}}
\newcommand{\Ymat}[0]{{{\boldsymbol Y}}}

\newcommand{\hv}[0]{{\boldsymbol{h}}}

\newcommand{\xv}{\boldsymbol{x}}
\newcommand{\yv}{\boldsymbol{y}}

\newcommand{\zv}{\boldsymbol{z}}

\newcommand{\Phimat}{\boldsymbol{\Phi}}

\newcommand{\tsp}{^{\mathsf{T}}}
\newcommand{\inv}{^{-1}}

\usepackage{newfloat}
\usepackage{listings}
\DeclareCaptionStyle{ruled}{labelfont=normalfont,labelsep=colon,strut=off} 
\floatstyle{ruled}
\newfloat{listing}{tb}{lst}{}
\floatname{listing}{Listing}
\title{Detail Matters: Mamba-Inspired Joint Unfolding Network for \\Snapshot Spectral Compressive Imaging}
\author{
    Mengjie Qin\textsuperscript{\rm 1,2}, Yuchao Feng\textsuperscript{\rm 1,2}, Zongliang Wu\textsuperscript{\rm 1}, Yulun Zhang\textsuperscript{\rm 3}, Xin Yuan\textsuperscript{\rm 1}\thanks{Corresponding author. (xyuan@westlake.edu.cn)}\\
}
\affiliations{
    \textsuperscript{\rm 1}School of Engineering, Westlake University, Hangzhou, China.\\
    \textsuperscript{\rm 2}Westlake Institute for Optoelectronics, Fuyang, Hangzhou, Zhejiang 311421, China.\\
    \textsuperscript{\rm 3}Shanghai Jiao Tong University, Shanghai, China.\\

}

\usepackage{bibentry}

\begin{document}

\maketitle

\begin{abstract}
In the coded aperture snapshot spectral imaging system, Deep Unfolding Networks (DUNs) have made impressive progress in recovering 3D hyperspectral images (HSIs) from a single 2D measurement. 
However, the inherent nonlinear and ill-posed characteristics of HSI reconstruction still pose challenges to existing methods in terms of accuracy and stability.
To address this issue, we propose a Mamba-inspired Joint Unfolding Network (MiJUN), which integrates physics-embedded DUNs with learning-based HSI imaging.
Firstly, leveraging the concept of trapezoid discretization to expand the representation space of unfolding networks, we introduce an accelerated unfolding network scheme. This approach can be interpreted as a generalized accelerated half-quadratic splitting with a second-order differential equation, which reduces the reliance on initial optimization stages and addresses challenges related to long-range interactions.
Crucially, within the Mamba framework, we restructure the Mamba-inspired global-to-local attention mechanism by incorporating a selective state space model and an attention mechanism.
This effectively {\bf reinterprets Mamba as a variant of the Transformer} architecture, improving its adaptability and efficiency.
Furthermore, we refine the scanning strategy with Mamba by {\bf integrating the tensor mode-$k$ unfolding into the Mamba} network. This approach emphasizes the low-rank properties of tensors along various modes, while conveniently facilitating 12 scanning directions. 
Numerical and visual comparisons on both simulation and real datasets demonstrate the superiority of our proposed MiJUN, and achieving overwhelming detail representation. 
\begin{links}
    \link{Code}{https://github.com/Mengjie-s/MiJUN.}
\end{links}
\end{abstract}

%
\section{Introduction}
\begin{figure}[htbp!]
  \begin{center} 
   \includegraphics[width=\linewidth]{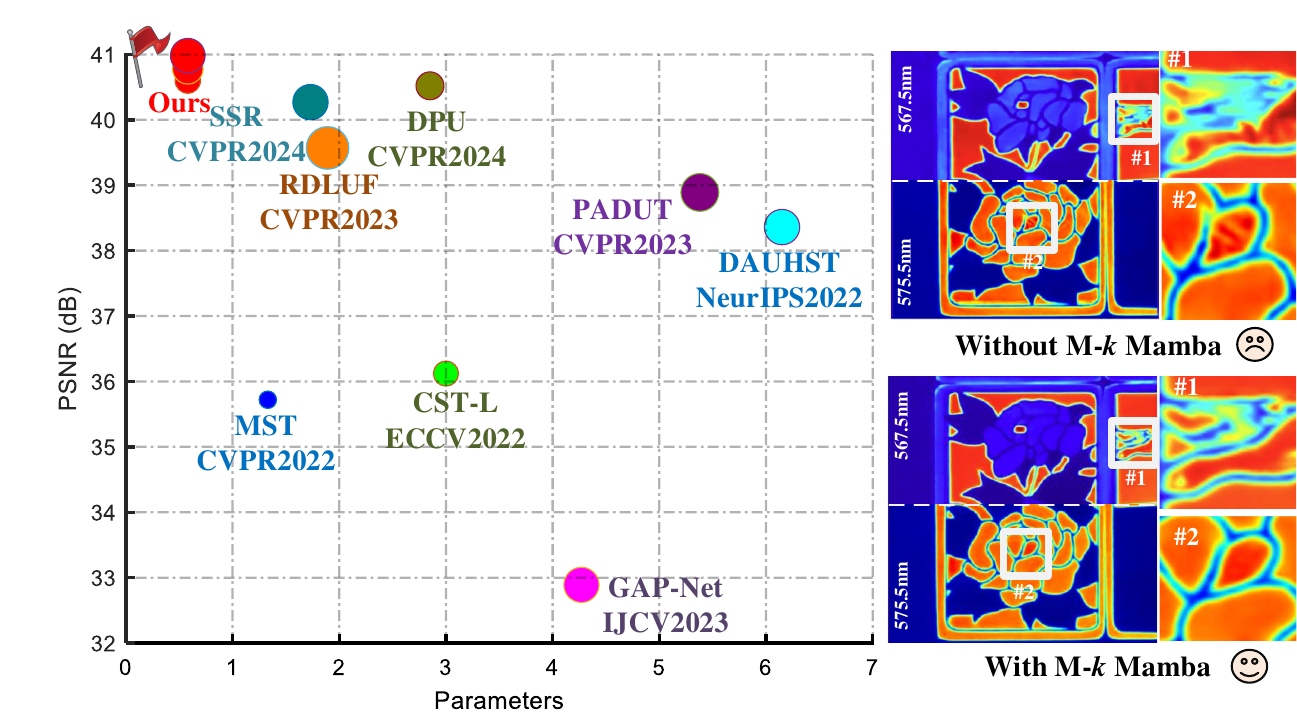}
  \caption{Comparison of reconstruction quality vs. Parameters(M), and FLOPs(G). Our proposed method outperforms comparisons, while utilizing less computational costs. Notably, the images on the right show the feature maps of RDULF and our method, where our features exhibit reduced noise and sharper edges.}\label{fig:para_psnr}
  \end{center}
\end{figure}
Coded Aperture Snapshot Spectral Imaging (CASSI) has emerged as a widely developed and utilized method for hyperspectral imaging.
This method is characterized by low bandwidth, rapid acquisition, and high throughput.
Technically, the CASSI process can be divided into two distinct phases. Initially, the 3D hyperspectral image (HSI) is encoded into a single 2D compressed measurement. Subsequently, the computational reconstruction phase employs reconstruction algorithms to estimate the original HSI from the snapshot measurement. This phase is the critical component of the entire CASSI system; therefore, developing high-quality reconstruction algorithms is imperative for the practical implementation of CASSI systems.

To address this challenge,  traditional model-based methods~\cite{bioucas2007new,liu2018rank,chen2023combining,luo2022hlrtf} often utilize regularization based on image priors to facilitate reconstruction.
Although these methods are highly interpretable, they are limited by their reliance on hand-crafted priors, which may lead to suboptimal results.
Recently, numerous deep learning-based approaches for CASSI have been developed. Based on differences in network structure, these approaches are generally divided into three categories: end-to-end methods~\cite{huang2021deep,meng2020end,cheng2022recurrent,meng2020end}, plug-and-play methods~\cite{chan2016plug,yuan2020plug,yuan2021plug,ebner2024plug}, and deep unfolding methods~\cite{wang2020dnu,wu2025latent,ma2019deep,zhang2022herosnet}.
The end-to-end (E2E) method typically constructs a direct mapping from the compressed measurement space to the original image domain. This approach significantly reduces computational complexity and often outperforms traditional model-based methods in terms of efficiency and effectiveness.
The plug-and-play (PnP) method incorporates a fixed pre-trained denoiser into traditional model-based frameworks without additional training. This integration employs pre-trained denoisers, which may not effectively adapt to the specific mappings required by different datasets.
Deep unfolding networks (DUNs) reconfigure specific optimization techniques into deep neural architectures. 
Specifically, DUNs endeavor to construct interpretable deep neural networks by integrating the framework of conventional iterative algorithms.
In this work, we focus on DUNs, which have been empirically proven to be successful in resolving optimization challenges.

Typically, DUNs integrate advanced network modules as denoisers to achieve robust interpretability and superior reconstruction capabilities. 
However, their performance remains uncertain due to reliance on approximated prior settings or insufficient feature learning.
Current unfolding algorithms often capture extensive dependencies by leveraging the Transformer framework.
Despite these algorithms achieving good results in existing HSI reconstruction tasks, they are still limited by the following issues:
(i) These models are developed based on Transformer networks, which have a very high computational cost, as the complexity of the attention being $\mathcal{O}(N^2)$.
(ii) There exists a trade-off between computational complexity and effective receptive field, which hinders these methods from exploring long-range dependencies, especially in HSIs.
Naturally, this prompts a compelling research question:
How can we design an HSI image reconstruction module to achieve a good balance between high performance and low model complexity?

Recently, the state space model (SSM) is a promising backbone for addressing the limitations of Transformers and CNNs.
The visualization Mamba model introduces a cross-scanning module, which applies the structured state space sequence (S4) model to visual tasks by unfolding 2D features into 1D arrays along four directions. This allows it to capture long-range context using a global receptive field with $\mathcal{O}(N)$ complexity.
However, as the Mamba model unfolds 2D features into 1D sequences, spatially adjacent pixels can become distant in the flattened sequence. This increased separation between neighboring pixels leads to a neglect of local context, resulting in a significant loss of essential local textures, thereby degrading HSI reconstruction performance.
To address the aforementioned issues, we propose a Mamba-inspired Joint Unfolding Network (MiJUN) for HSI reconstruction. 
Specifically, inspired by Mamba, we reformulate the SSM and the attention mechanism in a unified framework, describing Mamba as a variant of the Transformer, thereby leveraging the strengths of both Mamba and Transformer.
Furthermore, to address the issue of insufficient spatial and spectral feature representation in HSIs, we are the first to integrate the tensor mode-$k$ unfolding strategy into Mamba.
Finally, we introduce an acceleration strategy-based HQS (A-HQS), which can be regarded as a second-order differential equation, featuring improved convex approximation and $\mathcal{O}(1/k^2)$ convergence rates, while the first-order convergence rate is $\mathcal{O}(1/k)$.
As shown in Fig.~\ref{fig:para_psnr}, our MiJUN-5stg outperforms the previous SOTA RDLUF-MixS$^2$-9stg~\cite{dong2023residual} by 1.01 dB in PSNR value, with 3$\times$ fewer parameters and 3$\times$ less computational cost.

In summary, we present a {\bf joint unfolding network} for spectral SCI reconstruction, which integrates {\em mode-$k$ tensor unfolding} into the Mamba framework and then feeds into the accelerated {\em deep unfolding} network.
The principal contributions are as follows:
\begin{itemize}
  \item We propose a Mamba-inspired accelerated unfolding network for compressive spectral snapshot imaging, which formulates Mamba and Transformer in a unified framework.
  It retains the inherent advantages of the Mamba,  while achieving global-to-local information complementation through the attention module.
  
  \item Mode-$k$ tensor unfolding is first incorporated into the Mamba module, which reduces complex tensor operations to relatively easy-to-handle matrix operations by unfolding 3D tensors along each mode. This bridges the high-dimensional input form and the vector form required by Mamba, while conveniently emphasizing low-rankness and achieving 12-direction scanning. 
  
  \item We introduce an interpretable A-HQS for the solution of the DUN model. Based on this iterative solution framework, redundant elements can be effectively discarded, thereby accelerating the convergence of iterations. 
   \item The comprehensive evaluation conducted on both simulated and real datasets confirms that our proposed method exhibits superior quantitative performance, enhances visual quality, and reduces computational demands. Moreover, it excels at recovering fine details in the image. 
\end{itemize}

\section{Related Work}
\subsection{Vision Transformer for CASSI}
Previous studies have employed end-to-end neural networks to develop data-driven priors, which have been extensively applied in SCI applications.
Recent related researches~\cite{hu2022hdnet,zhang2021attention} have also confirmed that CNN-based methods exhibit strong capabilities to model local similarities.
However, despite their strengths, CNN-based techniques are constrained by their inductive biases, limiting their ability to identify non-local similarities.

To address the aforementioned issues, Transformer-based approaches~\cite{luo2024dual,wang2022spatial,cao2024hybrid} have gained significant popularity in computer vision due to their exceptional ability to model long-range interactions across spatial regions. 
However, these algorithms exhibit deficiencies in capturing the local features of HSI, failing to adequately represent the detailed and textured information of the images. ~\cite{cai2022mask} employ multi-head self-attention (MSA) mechanisms to capture long-range spatial and spectral dependencies in HSI. Using MSA, it computes the spectral dependencies, resulting in an attention map that implicitly encodes the global context.
Moreover,~\cite{cai2022degradation} introduce a half-shuffle MSA mechanism, which divides attention heads into a local branch and a non-local branch. 
This method models non-local similarity by shuffling pixels, which brings distant pixels into a local window. However, this technique can only capture non-local similarities of specific pixels, potentially overlooking highly correlated non-local pixels. Additionally, focusing on pixel-level non-local similarities may miss some object-level non-local similarities. 
Therefore, designing a network that effectively leverages both local and patch-level non-local priors in HSIs is of great importance.

\subsection{State Space Model}
SSMs were originally developed as a mathematical framework to describe system dynamics in motion. ~\cite{gu2021combining} introduced the linear State-Space Layer, combining the strengths of recurrent neural networks, temporal convolutions, and neural differential equations to improve model capacity. Building on this, some models ~\cite{gu2021efficiently, xie2024fusionmamba} leverage the optimization of SSM to address the issue of long-range dependencies, significantly improving computational efficiency.


Recently, SSM has gained increasing attention, being widely applied in natural language processing and gradually extending to visual tasks.
Intuitively, Mamba~\cite{gu2023mamba} is a state-space model that varies over time based on a gating mechanism, which effectively captures long-sequence dependencies. 
~\cite{liu2024vmambavisualstatespace} introduce a general visual backbone, Vim, which integrates bidirectional Mamba modules. This approach leverages positional embeddings to encode image sequences and employs a bidirectional state-space model (SSM) to compress visual representations.
In ~\cite{pei2024efficientvmamba} improves efficiency with a redesigned selective scanning method. However, directly applying Mamba to HSI reconstruction faces challenges, including loss of local context and key textures.

\section{Methodology}
\subsection{Degradation model of CASSI}
\begin{figure}[!htb]
  \begin{center}
   \includegraphics[height=2.5cm]{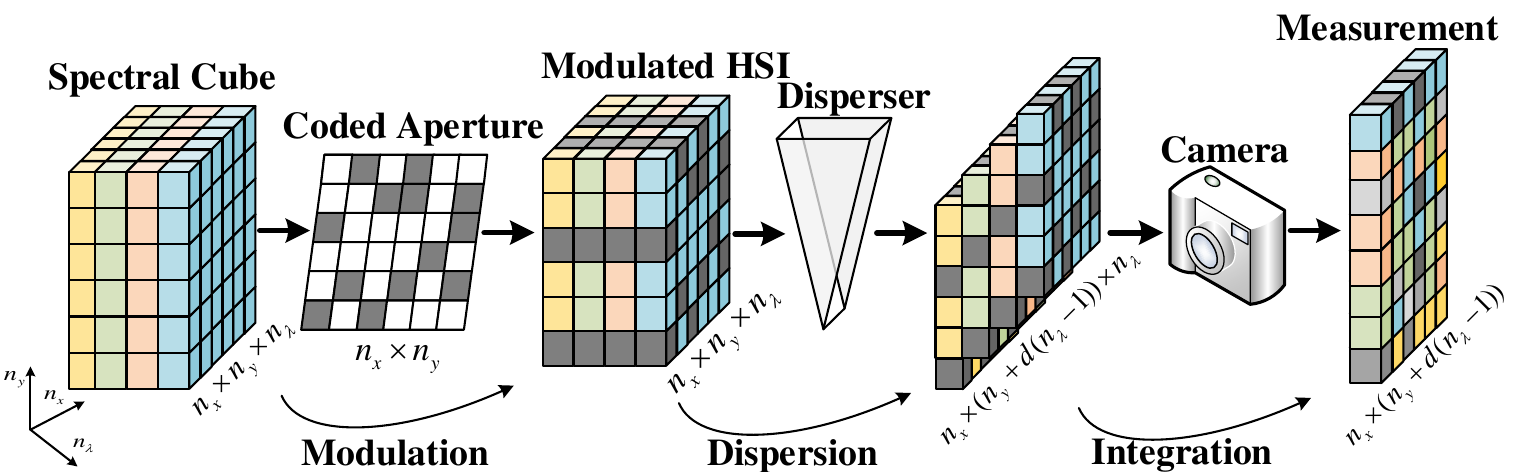}
  \caption{ A schematic diagram of CASSI.}\label{fig:cassi}
  \end{center}
\end{figure}

In the CASSI system, the 3D HSI cube is modulated by a physical mask in the aperture and incorporated different wavelengths through 2D monochrome sensors along the width dimension, ultimately compressed into a single 2D measurement. 
Fig.~\ref{fig:cassi} illustrates the forward imaging process of the single-disperser CASSI (SD-CASSI).   Mathematically, the original HSI data is denoted as 
$\Xmat\in\mathbb{R}^{W\times H\times N_{\lambda}}$,  $W$ and $H$ are the spatial dimensions, and $N_{\lambda}$ is the number of spectral channels. Similarly, the physical mask is denoted $\bm{M}\in\mathbb{R}^{W\times H}$. 
The coded HSI data cube at $n_{\lambda}$-th wavelength is represented as $\Xmat^{'}_{n_{\lambda}}=\Xmat_{n_{\lambda}}\odot\bm{M}$, where $\odot$ is the element-wise multiplication.     

Subsequently, passing the disperser, the spatially modulated HSI $\Xmat^{'}$ is tilted along the $H$-axis, which can be formulated as $\Xmat^{''}\in \mathbb{R}^{W\times \tilde{H}\times N_{\lambda}}$ and $\tilde{H}=H+d_{N_{\lambda}}$. In this context, $d_{N_{\lambda}}$ denotes the displacement magnitude experienced by the wavelength corresponding to the $N_{\lambda}$-the order. 
This operation can be formally described as the modulation of the shifted spectral component, denoted as 
$\tilde{\Xmat} \in \mathbb{R}^{W \times \tilde{H} \times N_{\lambda}}$, by employing a correspondingly shifted mask $\tilde{\bm{M}} \in \mathbb{R}^{W\times \tilde{H} \times N_{\lambda}}$. 
The relation for $\tilde{\bm{M}}$ at any given position is articulated as $\tilde{\bm{M}}(i, j, n_{\lambda})= \bm{M}(w, h + d_{\lambda})$. Finally, the imaging sensor acquires the dispersed as a 2D measurement $\bm{Y}$ can be formulated as follows:
%
\begin{equation}
   \Ymat= \textstyle \sum_{n=1}^{N_{\lambda}} \tilde{\Xmat}(:,:,n_{\lambda}) \odot \tilde{\bm{M}}(:,:,n_{\lambda}) + \Bmat,
 \label{eq1}
\end{equation}
where $\Bmat$ denotes the additive noise. Mathematically, by vectorizing $\Xmat$ and $\bm{Y}$, the above equation can be formulated as:
\begin{equation}
   \yv=\Phimat\xv+\bm{b},
\label{eq2}
\end{equation} 
where $\xv\in \mathbb{R}^{W\tilde{H}N_{\lambda}}$, $\yv\in \mathbb{R}^{W\tilde{H}}$. Here, $\Phimat\in \mathbb{R}^{W\tilde{H}\times W\tilde{H}N_{\lambda}}$ denotes the sensing matrix, which is generally construed as the spatially shifted mask within the imaging apparatus. 
\begin{figure*}[!htb]
  \begin{center}
   \includegraphics[height=4.2cm]{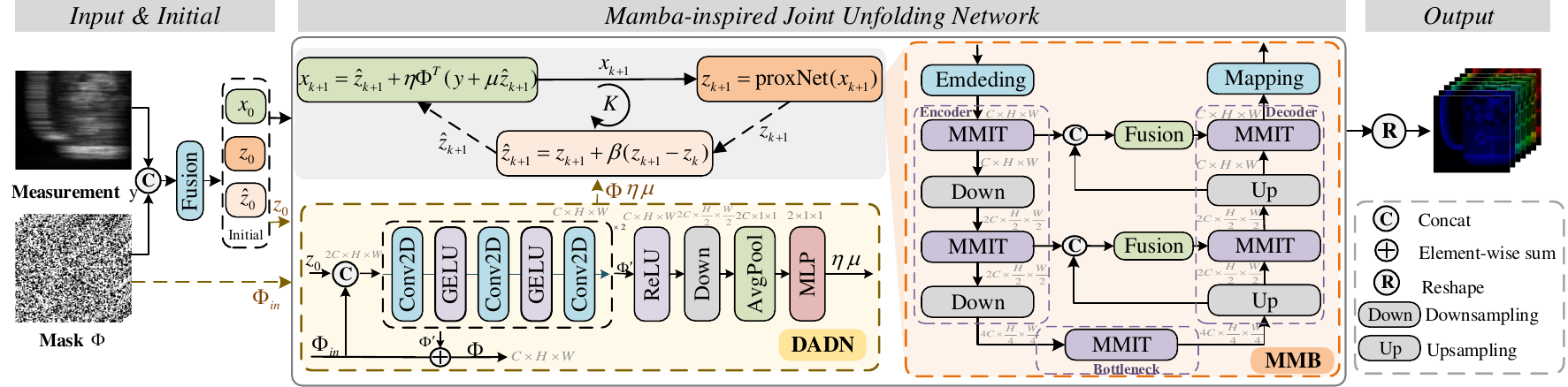}
  \caption{An overview of our proposed MiJUN for HSI reconstruction task, including input \& Initial, MiJUN model, and  Output. 
The model includes three iterative operators: $\xv$, $\zv$, $\hat{\zv}$. During the iterative process, the parameters are estimated by the DADN, with $\hat{\zv}$ learned through the MMB block. }\label{fig:Flow_chart}
  \end{center}
\end{figure*}
\begin{figure*}[!htb]
  \begin{center}
   \includegraphics[height=4.7cm]{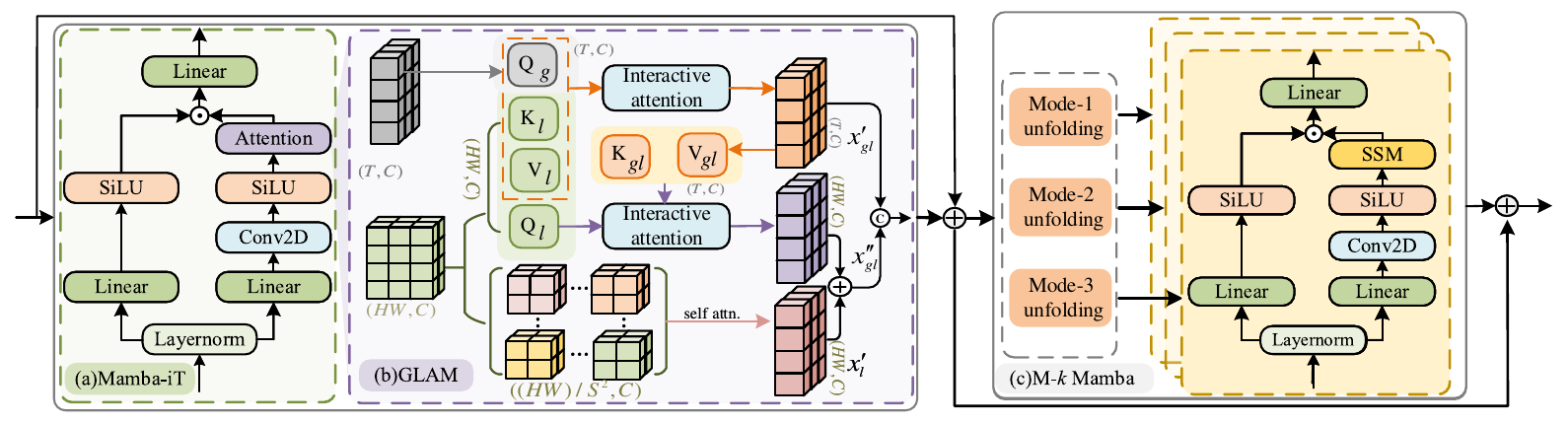}
  \caption{The diagram of the proposed MMIT. Features are first sufficiently modeled with local and global information through the Mamba-i T module, followed by the M-$k$ Mamba to further enhance the low-rank attributes.}\label{fig:MMIT}
  \end{center}
\end{figure*}

\subsection{Accelerated deep unfolding framework}
Recall that directly inferring the HSI $\xv$ from the degradation model Eq.~(\ref{eq2}) is intractable. 
Therefore, it is necessary to utilize the regularizer to constrain the solution space, and the inversion of the Eq.~(\ref{eq2}) can be construed as an optimization effort aimed at minimizing the cost function:
\begin{equation}
  \textstyle  \argmin_{\xv}\frac{1}{2}\|\yv-\Phimat\xv\|^2+\tau \mathcal{R}(\xv),
   \label{eq3}
\end{equation} 
where $\mathcal{R}(\xv)$ is the regularization term, characterizing the prior knowledge of the desired $\xv$, and $\tau$ denotes the noise-balancing factor. 
Within the research of SCI, previous deep unfolding models disregarded the implications of {\em accelerated} optimization algorithms, thereby failing to exploit {\em second-order gradient} information adequately.
We augment the HQS algorithm, previously used in DUN, to bridge these two aspects with an improved accelerated variant.
For clarity, we hereby briefly describe the iterative frameworks of $\mathit{A}$-HQS to solve the Eq.~\eqref{eq3}, which proceeds as follows:
%
\begin{subequations}
\begin{align}
   &\xv_{k+1}= \textstyle \argmin_{\xv}\frac{1}{2}\|\yv-\Phimat\xv\|^2+\frac{\mu}{2}\|\xv-\hat{\zv}_{k+1}\|^2,\\
   &\zv_{k+1}= \textstyle \argmin_{\zv}\frac{\mu}{2}\|\xv_{k+1}-\zv_k\|^2+ \tau \mathcal{R}(\zv_k),\\
   &\hat{\zv}_{k+1}=\zv_{k+1}+\beta_{k+1}(\zv_{k+1}-\zv_{k}), 
\end{align}\label{eq4}%
\end{subequations} where $\beta$ is the balancing parameter. 
For the subproblem $\xv_{k+1}$, it should be noted that Eq.~(\ref{eq4}a) is differentiable and the gradient descent scheme can be borrowed as a solver:
\begin{equation}
  \begin{split}
   \xv_{k+1}=(\Phimat{\tsp}\Phimat+\mu \Imat)\inv (\Phimat{\tsp}\yv+\mu\hat{\zv}_{k}),
  \end{split}\label{eq5}
\end{equation} 
where $\Imat$ denotes the identity matrix with desired dimensions. The matrix can be regarded as a regularized version (by adding $\mu\Imat$) of the Hessian of $\frac{1}{2}\|\yv-\Phimat\xv\|^2$. Therefore, the optimization process also manifests the application of second-order information in this aspect. 
In CASSI systems, $\Phimat$ is a fat matrix,  and $(\Phimat{\tsp}\Phimat+\mu\Imat)$ form a large matrix. Therefore, based on the Sherman-Morrison-Woodbury formula, the equation can be simplified to the following:
\begin{equation}
  \begin{split}
   \xv_{k+1}&=[\mu^{-1} \Imat-\mu^{-1}\Phimat{\tsp}(\Imat+\Phimat\mu^{-1}\Phimat{\tsp})^{-1}\Phimat\mu^{-1}]\\
   &~~\times [\Phi{\tsp}\yv+\mu\hat{\zv}_{k}].
  \end{split}\label{eq6}
\end{equation} 
For SCI in this paper, $\Phimat\Phimat{\tsp}$ corresponds to an identity matrix interspersed with zeros on its diagonal (matching the locations of the undetermined observations) as:
\begin{equation}
  \begin{split}
  \Phimat\Phimat{\tsp}=\mathrm{diag}\{r_1,\cdots,r_N\},
  \end{split}\label{eq7}
\end{equation}  
where $N$ represents the number of rows in $\Phimat$. 

Consequently, Eq.~(\ref{eq5}) simply involves multiplication $(\Phimat{\tsp}\yv+\mu\hat{\zv}_{k})$ by $\Phimat{\tsp}\Phimat$, which is an operation with $\mathcal{O}(n)$.  Eq.~(\ref{eq5}) can be simplified as:
\begin{equation}
\xv_{k+1}=\hat{\zv}_{k}+\Phimat{\tsp}(\yv-\Phimat\hat{\zv}_{k})\oslash[\mu+\mathrm{diag}(\Phimat\Phimat{\tsp})],
\label{eq8}
\end{equation} 
where $\oslash$ is the element-wise division of Hadamard division.

For the $\zv$-subproblem, Eq.~(\ref{eq4}b) is a deterministic approximation operator predicated on a specified prior $\mathcal{R}(\zv)$. Unfortunately, the inherent uncertainty associated with the function $\mathcal{R}(\zv)$ precludes the availability of any closed-form solutions.
Thus, we propose an Mamba-inspired and Mamba model to function as the prior extractor, which can generally be formulated as:
\begin{equation}
   \zv_{k+1}=\mathrm{proxNet}_{\tau,\mu}(\xv_{k+1}).
\label{eq9}
\end{equation} 
Additionally, $\mathrm{proxNet}_{\tau,\mu}$ can be represented as a denoiser $\mathcal{D}_{\eta}$ with learnable noise level $\eta$.
The overall iterative MiJUN framework is shown in Fig.~\ref{fig:Flow_chart}. And, we introduce the Mamaba and Mamba-inspired Block (MMB) to play the role of $\mathrm{proxNet}$, which features a U-shaped network and mainly includes a submodule of Mamba and Mamba-inspired Transformer (MMIT).
The detailed description is as follows.

\begin{figure*}[htbp!]
  \begin{center} 
  \includegraphics[width=14cm,height=4.8cm]{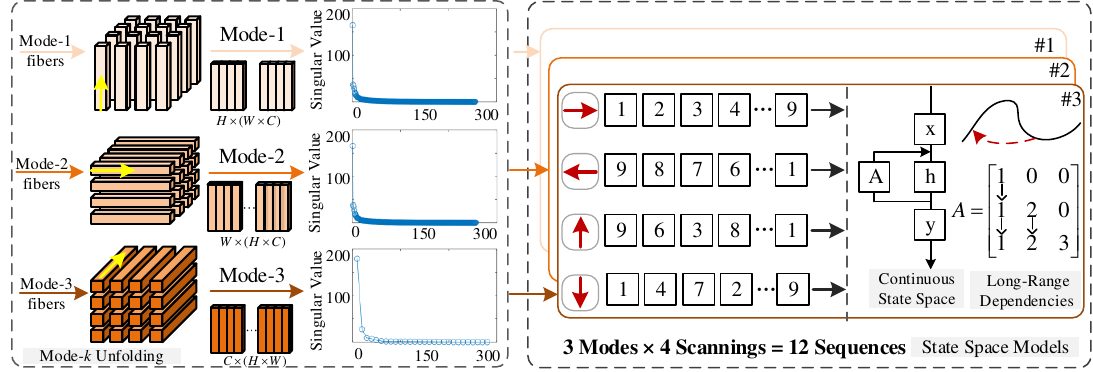}
  \caption{Illustration of Mode-$k$ unfolding along each direction of 3D tensor and  linear-overhead SSM with different-direction scanning scheme. The low rank of each matrix after unfolding is demonstrated by singular value decomposition (SVD(log)).}\label{fig:modek-ssm}
  \end{center}
\end{figure*}

\subsection{Prior extractor}
\noindent{\bf Mamba-inspired Transformer.}
Empirically, Mamba has been shown to perform well in tasks requiring global context understanding. 
However, despite these strengths, Mamba encounters challenges in adequately representing local texture features. This inadequacy arises because the linear unfolding of 2D features into 1D sequences can lead to the loss of spatially adjacent pixel relationships, crucial for capturing fine-grained local details. The large distance between neighboring pixels in the flattened sequence can result in a neglect of local context, leading to a significant loss of key local textures and reduced performance in tasks that require detailed local feature extraction.

By theoretically and empirically analyzing Mamba from the perspective of linear attention Transformer~\cite{han2024demystify}, and integrating strategies to enhance local feature extraction within the Mamba framework can potentially address this limitation and improve the performance of HSIs reconstruction. 
Specifically, we reformulate selective SSM and attention within a unified framework, describing the Mamba-inspired Transformer (Mamba-i T) as a variant in Fig.~\ref{fig:MMIT}(a). 
Following~\cite{liu2024vmambavisualstatespace}, the input feature $\Xmat \in \mathbb{R}^{W\times H \times C}$  is processed through two parallel branches. 
One branch consists of channel expansion, a linear layer, and SiLU activation.
In the other branch, the channels are first expanded to $\lambda C$ using a linear layer, where $\lambda$ is a predefined channel expansion factor. 
Then, features are extracted using a $3 \times 3$ convolution followed by SiLU activation. Finally, these features are processed through an attention mechanism.  
Notably, we adopt a global-local attention mechanism (GLAM) to compensate for Mamba's deficiencies in capturing spatial local features shown in Fig. \ref{fig:MMIT}(b).
After aggregating the features of both branches, the channels are projected back to
$C$, producing an output $\Xmat_{out}$, as follows.
%
\begin{align}
   &\mathrm{Branch 1}: \Xmat_1=\mathrm{SiLU}(\mathrm{Lin}(\mathrm{Ln}(\Xmat))),   \nonumber\\
   &\mathrm{Branch 2}: \Xmat_2=\mathrm{GLAM}(\mathrm{SiLU}(\mathrm{Conv}(\mathrm{Lin}(\mathrm{Ln}(\Xmat))))),   \nonumber\\
    &\mathrm{Output}: \Xmat_{out}=\mathrm{Lin}(\Xmat_1\odot \Xmat_2),
  \nonumber
\end{align} 
where $\mathrm{Ln}$ is layernorm, $\mathrm{Lin}$  represents the linear layer, and $\odot$ is Hadamard product.

\noindent{\bf Mode-$k$ unfolding-based Mamba.}
Considering the spatial complexity and spectral similarity of HSIs, we adopt a tensor mode-$k$ unfolding strategy to capture both spatial and spectral structures. This approach preserves essential spatial features that might otherwise be lost with traditional channel-slicing methods. As illustrated in Fig.~\ref{fig:modek-ssm} with Scene 1, the schematic shows the image for each mode-$k$ unfolding matrix and their corresponding singular values. Notably, Mode-1 and Mode-2 share similar singular value distributions, while Mode-3 exhibits a distinct pattern.


%
\begin{figure*}[!htb]
  \begin{center}
   \includegraphics[width=17.6cm]{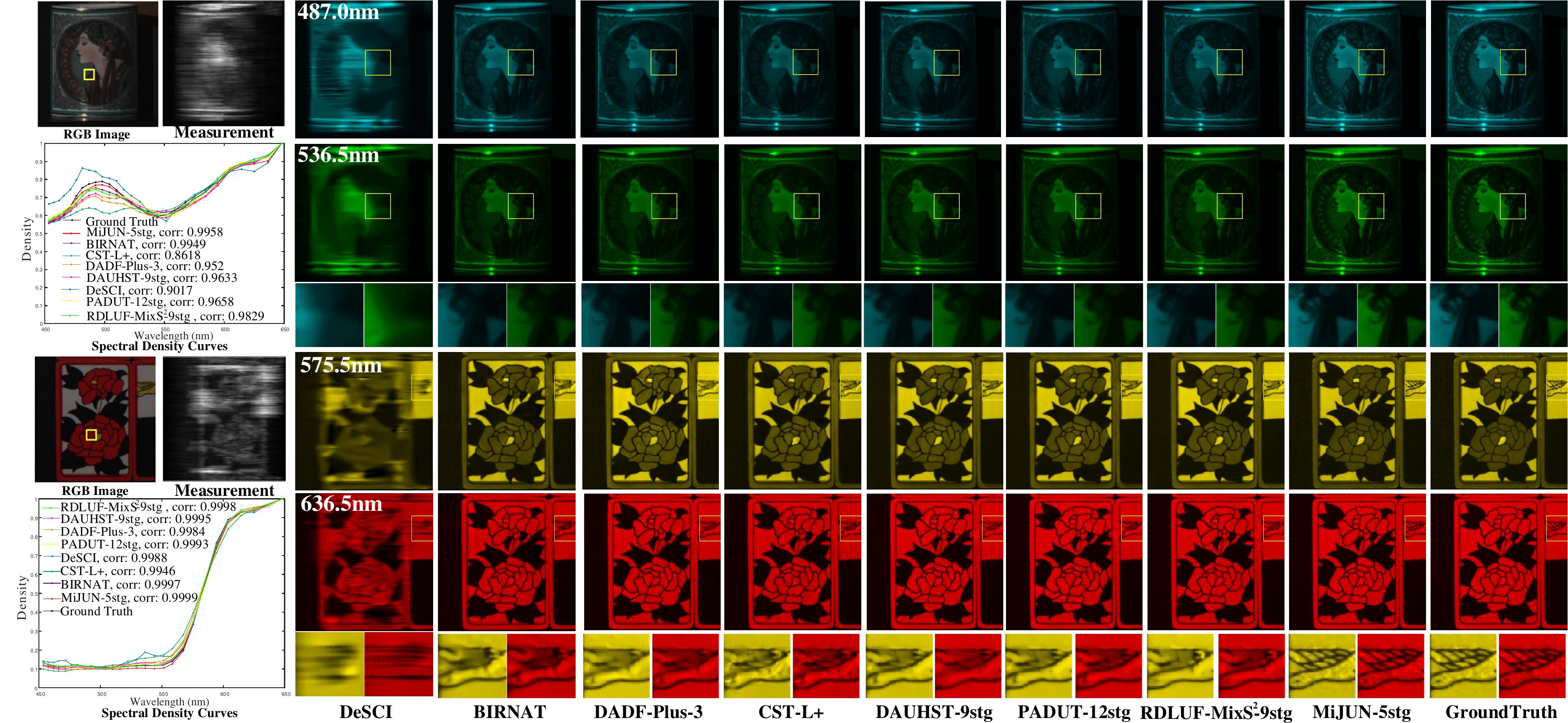}
  \caption{The simulated HSI reconstruction results for Scene 1 (top) \& Scene 7 (bottom) with 2 out of 28 spectral channels, including seven state-of-the-art algorithms and our proposed MiJUN-5stg. The left displays the RGB image and measurement. The bottom-left shows the spectral density curves corresponding to the selected yellow box in the RGB image. }\label{fig:vision-sim}
  \end{center}
\end{figure*}

Therefore, hereby we integrate the tensor mode-$k$ unfolding strategy into the Mamba network, 
proposing the Mode-$k$ unfolding-based Mamba (M-$k$ Mamba), which is illustrated in the left of Fig.~\ref{fig:modek-ssm}.
First, the input data undergoes a tensor mode-$k$ unfolding transformation to obtain different tensor unfolded data, {\em i.e.}, $\Xmat\in \mathbb{R}^{B\times WH \times C}$.
Then, the data is fed into the model following the previously mentioned Mamba network~\cite{liu2024vmambavisualstatespace}.
The difference lies in Branch 2, where $\Xmat_2$ is computed using the SSM.
By structuring the Mamba input to depend on the long-range representation of parameter ‘$\Amat$’, it effectively filters out irrelevant information, allowing more efficient compression of the context into the hidden state. As shown in Fig.~\ref{fig:modek-ssm}, in the SSM, the given input is unfolded into four one-dimensional sequences/vectors $\{\xv_n \in \mathbb{R}^{1\times \hat{H}\hat{W}\hat{C}}\}_{n=1}^4$ by scanning pixels along four different traversal paths: from top-left to bottom-right, from top-right to bottom-left, from bottom-right to top-left, and from bottom-left to top-right. Notably, combining 3 modes of mode-$k$, 12-direction scanning sequences can be conveniently derived.
Subsequently, SSM is calculated as follows.
\begin{equation}
  \begin{split}
   &\{\Bmat_t, \Delta_t, \Cmat_t\}={\cal P}_{\rm proj}(\xv_n), \quad \overline{\Delta_t}= \sigma ^+({\cal P}_{\rm dt}\Delta_t),\\
   &\overline{\Amat_t}=\exp(-\exp(\Amat_{\log}\overline{\Delta_t})), \quad \overline{\Bmat_t}=\overline{\Delta_t} \odot \Bmat_t,
  \end{split}\label{eq11}
    \nonumber
\end{equation} 
where ${\cal P}_{\rm proj}$, ${\cal P}_{\rm dt}$, and $A_{\rm log}$ are time-invariant weight matrices, $\sigma ^+$ is softplus activation function, and $\odot$ is element-wise multiplication. 
Weight matrices $\Bmat$ and $\Cmat$ directly depend on input $\xv_n$, whereas recurrent weight matrix $\Amat$ depends solely on the time-scale parameter $\Delta$. 
The hidden state $\hv$ and output $\yv$ of SSM are calculated as follows.
\begin{equation}
\hv_t=\overline{\Amat} \odot\hv_{t-1}+\overline{\Bmat} \odot \xv_{n}, \quad \yv_t=\Cmat_t \hv_t + \Dmat_t\odot \xv_n,
    \nonumber
\end{equation} 
where $\Dmat$ is the scale parameter, $._t$ represents the $t$-th state.

\section{Experiments}
\subsection{Experimental settings} 
\noindent{\bf Datasets.}
In the simulation experiments, we use two datasets, CAVE and KAIST. The CAVE dataset comprises 32 HSI images with spatial dimensions of $512 \times 512$. The KAIST dataset contains 30 HSI images, each with spatial dimensions of $2704 \times 3376$.
Same as previous researches, we utilize the CAVE dataset as our training set and selected 10 scenes from the KAIST dataset for testing.
In real experiments, we use five real CASSI datasets~\cite{meng2020end}, with dimensions of $660\times 714 \times 28$,  wavelength range from 450 to 650 nm and a dispersion of 54 pixels.

\noindent{\bf Implementation Details.}
The proposed model MiJUN is implemented by Pytorch.
During the training process, we utilize the Adam optimizer ($\beta_1=0.9$ and $\beta_2=0.999$) and a cosine annealing scheduler, running for 200 epochs on a single RTX 4090 GPU.
To evaluate the performance, we use the peak signal-to-noise ratio (PSNR) and structure similarity (SSIM) to assess the HSI reconstruction capabilities.

\subsection{Compare with State-of-the-art}
We compare our proposed MiJUN model with several SOTA CASSI algorithms and the results are analyzed as follows. 

\begin{table*}[ht!]
\begin{center}
  \resizebox{0.98\textwidth}{!}
   {
  \begin{tabular}{c|cc|cccccccccc|c}
    \toprule
Algorithms&Params(M) &FLOPs(G) & Scene1 & Scene2 & Scene3 & Scene4 & Scene5 & Scene6 & Scene7 & Scene8 & Scene9 & Scene10 & Avg \\ \midrule
\rowcolor{gray!5}  &  & & 25.16 & 23.02& 21.40 & 30.19& 21.41 & 20.95 & 22.20 & 21.82                 & 22.42  & 22.67& 23.12\\
\rowcolor{gray!5} \multirow{-2}*{TwIST~\cite{bioucas2007new}}  & \multirow{-2}*{---}& \multirow{-2}*{---}& 0.700         & 0.604& 0.711& 0.851& 0.635 & 0.644& 0.643& 0.650& 0.690& 0.569& 0.669\\ 
\midrule

 \rowcolor{gray!5}  && & 26.82&22.89&26.31&30.65&23.64&21.85&23.76&21.98&22.63&23.1&24.36\\
\rowcolor{gray!5}   \multirow{-2}{*}{GAP-TV~\cite{yuan2016generalized}}   &  \multirow{-2}*{---} & \multirow{-2}*{---}   & 0.754&0.610&0.802&0.852&0.703&0.663&0.688&0.655&0.682&0.584&0.669\\ 
\midrule

 \rowcolor{gray!5}  && & 27.13 & 23.04  & 26.62& 34.96& 23.94& 22.38& 24.45 & 22.03& 24.56& 23.59& 25.27 \\
 \rowcolor{gray!5}  \multirow{-2}{*}{DeSCI~\cite{liu2018rank}}   &  \multirow{-2}*{---} & \multirow{-2}*{---}   & 0.748& 0.620 & 0.818& 0.897& 0.706& 0.683& 0.743  & 0.673& 0.732 & 0.587 &0.721  \\ 
\midrule
               
 \rowcolor{orange!5} &  &   & 34.96
&35.64&35.55&41.64&32.56&34.33&33.27&32.26&34.17&32.22&34.66 \\
\rowcolor{orange!5} \multirow{-2}{*}{HDNet~\cite{hu2022hdnet}}&\multirow{-2}*{2.37}  &  \multirow{-2}*{154.76}        & 0.937&0.943&0.94&0.976&0.948&0.95&0.92&0.945&0.944&0.94&0.946\\ 
\midrule

 \rowcolor{orange!5}  &  &  &36.78&37.89&40.61&46.93&35.42&35.30&36.58&33.95&39.46&32.80&37.57\\
\rowcolor{orange!5}    \multirow{-2}{*}{BIRNAT~\cite{cheng2022recurrent}}     & \multirow{-2}*{4.40} &  \multirow{-2}*{212.55}  &0.951&0.957&0.971&0.985&0.963&0.959&0.954&0.955&0.969&0.937&0.960\\ 
\midrule

 \rowcolor{orange!5}    &   &           
& 37.46& 39.86 & 41.03 & 45.98& 35.53& 37.02& 36.76& 34.78& 40.07& 34.39 & 38.29\\
 \rowcolor{orange!5}\multirow{-2}{*}{DADF-Plus-3~\cite{xu2023degradation}}  & \multirow{-2}*{58.13} & \multirow{-2}*{230.41} &0.965 &0.976 &0.974 &0.989 &0.972 &0.975 &0.958 &0.971 &0.976 &0.962 & 0.972   
\\ 
\midrule

 \rowcolor{orange!5}    & & & 35.57&36.22&37.00&42.86&33.27&35.27&34.05&33.50&36.17&33.26&35.72\\
  \rowcolor{orange!5} \multirow{-2}{*}{MST++~\cite{cai2022mask}}  &  \multirow{-2}*{1.33} & \multirow{-2}*{19.42}& 0.945
&0.949&0.959&0.980&0.954&0.960&0.936&0.956&0.956&0.949&0.955
                \\ 
\midrule

 \rowcolor{orange!5}&  &  & 35.96& 36.84& 38.16& 42.44& 33.25& 35.72& 34.86& 34.34 & 36.51 & 33.09& 36.12\\
  \rowcolor{orange!5}  \multirow{-2}{*}{CST-L+~\cite{cai2022coarse}}&\multirow{-2}*{3.00}  & \multirow{-2}*{40.10}           & 0.949& 0.955& 0.962& 0.975& 0.955& 0.963& 0.944& 0.961& 0.957& 0.945& 0.957 \\ 
\midrule 

  \rowcolor{green!5}   & && 31.72& 31.13 & 29.99& 35.34& 29.03  & 30.87 & 28.99  & 30.13& 31.03& 29.14& 30.74 \\
   \rowcolor{green!5}      \multirow{-2}{*}{DNU~\cite{wang2020dnu}} &  \multirow{-2}*{1.19} &\multirow{-2}*{163.48}& 0.863& 0.846& 0.845& 0.908 & 0.833 & 0.887  & 0.839& 0.885& 0.876& 0.849& 0.863 \\ 
\midrule

  \rowcolor{green!5}  && &  33.63
& 33.19& 33.96& 39.14& 31.44& 32.29& 31.79& 30.25& 33.06& 30.14& 32.89\\
  \rowcolor{green!5} \multirow{-2}{*}{GAP-Net~\cite{meng2023deep}}& \multirow{-2}*{4.27} &  \multirow{-2}*{78.58} &  0.913
& 0.902& 0.931& 0.971& 0.921& 0.927& 0.903& 0.907& 0.916& 0.898& 0.919
 \\ 
\midrule 

  \rowcolor{green!5}  & && 37.25& 39.02& 41.05& 46.15& 35.80& 37.08& 37.57 & 35.10& 40.02& 34.59& 38.36                  \\
   \rowcolor{green!5}  \multirow{-2}{*}{DAUHST-9stg~\cite{cai2022degradation}}   & \multirow{-2}*{6.15} &   \multirow{-2}*{79.50}       & 0.958& 0.967& 0.971& 0.983& 0.969& 0.970  & 0.963 & 0.966  & 0.970  & 0.956& 0.967  \\ 
\midrule

  \rowcolor{green!5} &  &  & 36.68 &38.74 &41.37 &45.79 &35.13 &36.37 &36.52 &34.40 &39.57 &33.78 &37.84                  \\
  \rowcolor{green!5} \multirow{-2}{*}{PADUT-5stg~\cite{li2023pixel}}   &  \multirow{-2}*{2.24} &  \multirow{-2}*{37.90}   & 0.955 &0.969 &0.975 &0.988 &0.967 &0.969 &0.959 &0.967 &0.971 &0.955 &0.967                  \\ 
\midrule

  \rowcolor{green!5}  &  &         &37.36 &40.43 &42.38 &46.62 &36.26 &37.27 &37.83 &35.33 &40.86 &34.55 &38.89                 \\
    \rowcolor{green!5}     \multirow{-2}{*}{PADUT-12stg~\cite{li2023pixel}}  &  \multirow{-2}*{5.38} &  \multirow{-2}*{90.46}   &  0.962 &0.978 &0.979 &0.990 &0.974 &0.974 &0.966 &0.974 &0.978 &0.963 &0.974                  \\ 
\midrule

   \rowcolor{green!5} &  & & 37.94& 40.95& 43.25 & 47.83   & 37.11 & 37.47& 38.58 & 35.50  & 41.83 & 35.23 & 39.57           \\
   \rowcolor{green!5} \multirow{-2}{*}{RDLUF-MixS$^2$-9stg~\cite{dong2023residual}}     & \multirow{-2}*{1.89} &  \multirow{-2}*{115.34}&0.966 & 0.977 & 0.979 & 0.990& 0.976  & 0.975 & 0.969  & 0.970 & 0.978  & 0.962  & 0.974           \\ 
\midrule

   \rowcolor{green!5}  & & & 38.52 &41.37 & 44.29 &\bf{48.84} &38.58 &38.08 &40.69 &\bf{36.93} &43.33 &35.41 & 40.60\\
  \rowcolor{green!5} \multirow{-2}{*}{MiJUN-5stg}     & \multirow{-2}*{0.56} &  \multirow{-2}*{40.98}     
&0.969&0.980&0.981&0.992&0.982&0.978&0.979&0.977&0.983&0.964&0.978   \\
\midrule

   \rowcolor{green!5} &           & 
& 39.10 &41.42&44.25 &48.78 &39.04 &37.97 &40.76 &36.46 &43.58 &\bf{35.64} &40.70\\
  \rowcolor{green!5} \multirow{-2}{*}{MiJUN-7stg}     & \multirow{-2}*{0.56} &  \multirow{-2}*{57.32}     
&0.971 &0.981 &0.981 &0.992 &0.983 &0.978 &0.979 &0.976 &0.984 &0.966 &0.979\\
\midrule

  \rowcolor{green!5}   &           & 
    &\bf{39.26}&\bf{41.78} &\bf{44.31} &48.53 &\bf{39.30} &\bf{38.22} &\bf{41.00} &36.72 &\bf{43.84} &35.56 &\bf{40.86} \\
  \rowcolor{green!5} \multirow{-2}{*}{MiJUN-9stg}     & \multirow{-2}*{0.56} &  \multirow{-2}*{73.67 }     
& \bf{0.973}&\bf{0.983}&\bf{0.983}&\bf{0.994}&\bf{0.985}&\bf{0.979}&\bf{0.983}&\bf{0.978}&\bf{0.985}&\bf{0.967}&\bf{0.982}   \\
\bottomrule
 \end{tabular}}
  \end{center}
\caption{The results of PSNR in dB (top entry in each cell), SSIM (bottom entry in each cell) on the 10 synthetic spectral scenes.`-5stg' denotes the network with 5 unfolding stages. `Avg'  represents the average of 10 scenes.}\label{Table:sim-grayscale}
\end{table*}

\noindent{\bf Synthetic data.}
To comprehensively evaluate the quantitative results of all competing methods, we test on ten simulated datasets and presented the corresponding numerical results in Tab. \ref{Table:sim-grayscale}. Different colors are used to distinguish the types of algorithms: gray for model-based methods, orange for end-to-end networks, and green for deep unfolding methods.
In Tab.~\ref{Table:sim-grayscale}, it is evident that our MiJUN model achieved the best numerical results in all cases.
Fig. \ref{fig:vision-sim} shows the visual reconstruction results. It is evident that MiJUN demonstrates a significant advantage over others, particularly in Scene 1, where it excels in detailing hair, and in Scene 7, where it effectively captures the intricacies of bird wings.
Furthermore, to evaluate overfitting, we test our pre-trained model on the unseen ICVL dataset \footnote{https://icvl.cs.bgu.ac.il/hyperspectral}, as shown in Tab.~\ref{Table:unseen_ICVL}, which demonstrates good generalization performance.

%
\begin{figure}[ht]
  \begin{center}
   \includegraphics[width=0.465\textwidth]{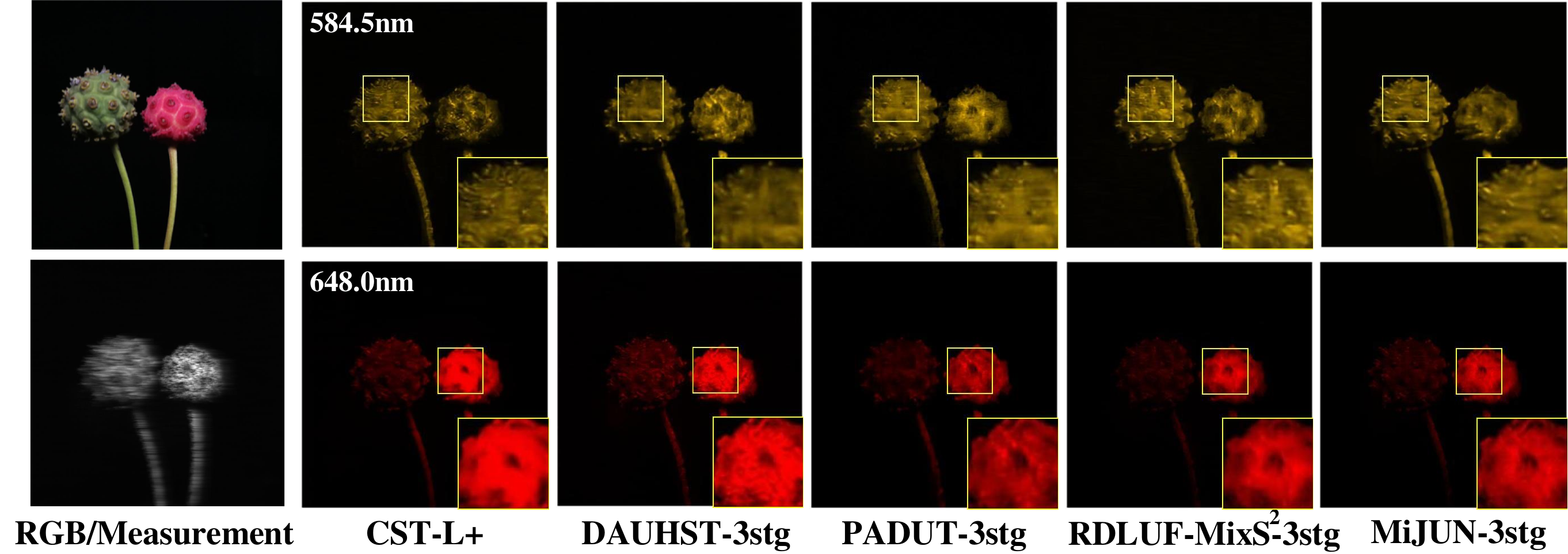}
  \end{center}
\caption{ The real data comparisons.  2 out of 28 wavelengths are plotted for visual comparison. }\label{fig:real}
\end{figure}

\noindent{\bf Real data.}
To further investigate the superiority of this model, we also conduct experiments on real HSI reconstruction tasks.
Since the ground truth of real-world scenarios is unattainable, we can only compare qualitative results.
Following the experimental setting of~\cite{cai2022mask}, we apply MiJUN-5stg to training in the simulated dataset.
Fig. \ref{fig:real}  presents the visual results of our model compared to other algorithms in Scene 4 (2 out of the 28 spectral channels).
In comparison, our model can reconstruct more textures and details, but it still exhibits some blurriness and artifacts. These challenges highlight the difficulties the model faces in handling real-world hyperspectral reconstruction tasks.

\begin{table}[h!]
\centering
\small 
\setlength{\tabcolsep}{1pt}
\begin{tabular}{lccccc}
\toprule
Method& eve\_0311 & BUG-0403& 4cam\_0411 & CC\_40D & guCAMP\_0514 \\
\midrule
RDLUF & 31.29 & 30.74 & 32.90 & 31.69 & 37.28 \\
Ours  & 31.97 & 31.27 & 33.82 & 32.18 & 37.78 \\
\bottomrule
\end{tabular}
\caption{Comparison of PSNR on the unseen ICVL dataset.}\label{Table:unseen_ICVL}
\end{table}

\begin{table}[htbp]
\setlength{\tabcolsep}{2.0pt}
\centering
\small

\begin{tabular}{cccccc}
\hline\hline
Methods&  PSNR $\uparrow$ & SSIM $\uparrow$\\ \hline
baseline& 38.59 & 0.969\\
$w/$Acc & 38.60& 0.971 \\
$w/$ GLAM&39.64 & 0.974\\ \hline\hline
\end{tabular}
\hfill
\begin{tabular}{cccccc}
\hline\hline
Methods&  PSNR $\uparrow$ & SSIM $\uparrow$\\ \hline
\raggedright $w/$Mamba-i&39.89 & 0.976\\
\raggedright $w/$Mamba&40.25 &0.976  \\
\raggedright $w/$M-$k$(ours)&\bf{40.60} &\bf{0.978}\\ \hline\hline
\end{tabular}
\caption{Ablation study of key components in our key components.
The $w/$ denotes the inclusion of a module. }\label{tab:ablation}
\end{table}

\subsection{Ablation study}
Our ablation analysis focuses on three main components of MiJUN, acceleration strategy, Mamba-i T and M-$k$ Mamba. We conduct ablation experiments on public simulated HSI datasets to investigate the effectiveness of each module. 
Tab.~\ref{tab:ablation} summarizes the performance of different cases compared to our model.
We select RDLUF-MixS$^2$-5stg, which combines the basic module, DADN, and the attention mechanism, as the baseline.
As shown in Tab.~\ref{tab:ablation}, when we incorporate the acceleration strategy into the baseline and replace the attention mechanism with GLAM, the model performance achieves a certain degree of improvement, with PSNR increasing from 38.59 to 39.64 and SSIM increasing from 0.969 to 0.974.
However, when we further integrate GLAM with the Mamba framework to validate the effectiveness of the Mamba-i T module(GLAB $\rightarrow$ Mamba-i T), we observe a 0.25 dB increase in PSNR.
Overall, the Mamba-i T module achieved an improvement of 1.3 dB in PSNR compared to the baseline.
Furthermore, we validate the effectiveness of integrating the tensor mode-$k$ unfolding with the Mamba network. In terms of PSNR results, when only the Mamba module is added, the PSNR increased to 40.25 dB.
Finally, by effectively integrating the tensor mode-$k$ unfolding with the Mamba module(Mamaba $\rightarrow$ M-$k$ Mamaba), we develop our complete model, MiJUN, which achieved the optimal result with a PSNR of 40.60 dB.
This ultimately confirms what is shown in Fig.~\ref{fig:para_psnr}: our model achieved the best results, with sharper image edges and richer details.


\section{Conclusion}
In this paper, we introduce a novel joint unfolding network for spectral snapshot compressive imaging, dubbed MiJUN. 
Firstly, based on the accelerated strategy, we construct an accelerated iteration scheme for DUN, enabling effective elimination of redundant information.
Additionally, inspired by Mamba, we incorporate a global-local attention mechanism into the Mamba framework as a variant of the Transformer architecture, effectively enhancing the model's feature extraction capabilities. Furthermore, to fully consider data characteristics, we introduce tensor mode-$k$ unfolding in the Mamba network, which enhances the representation of the intrinsic properties of the data.
This approach enables the model to learn features at a fine-grained level, facilitating detailed reconstruction of HSIs.
Comprehensive evaluations on both simulated and real datasets confirm the superior quantitative performance of our proposed approach. 

\section{Acknowledgments}
This work was supported by the National Key R$\&$D Program of China (grant number 2024YFF0505603, 2024YFF0505600), the National Natural Science Foundation of China (grant number 62271414), Zhejiang Provincial Outstanding Youth Science Foundation (grant number LR23F010001), Zhejiang “Pioneer" and “Leading Goose"R$\&$D Program(grant number 2024SDXHDX0006, 2024C03182), the Key Project of Westlake Institute for Optoelectronics (grant number 2023GD007), the 2023 International  Sci-tech Cooperation Projects under the purview of the “Innovation Yongjiang 2035” Key R$\&$D Program (grant number 2024Z126), and the Zhejiang Province Postdoctoral Research Excellence Funding Program (grant number ZJ2024086).

\bibliography{aaai25}

\end{document}